\documentclass{article}

\newcommand{\prob}[1]{\Pr\left[#1\right]}
\newcommand{\expect}[1]{E\left(#1\right)}
\newcommand{\var}[1]{\Var\left(#1\right)}
\newcommand{\tlim}{\lim_{t\rightarrow\infty}}

\usepackage{amsmath, amssymb,amsopn,amsthm} 
\usepackage{algorithmic, algorithm}   
\usepackage{verbatim}                 
\usepackage{xspace}                   
\usepackage{booktabs}                 

\usepackage{tikz,graphicx}
\usetikzlibrary{arrows,patterns,plotmarks,shapes,snakes,er,3d,automata,backgrounds,topaths,trees,petri,mindmap}

\DeclareMathOperator*{\argmin}{arg\,min}
\DeclareMathOperator*{\unif}{Unif}
\DeclareMathOperator*{\Var}{Var}

\newtheorem{theorem}       {{\bf Theorem}}
\newtheorem{lemma}         {{\bf Lemma}}

\newtheorem{proposition}   {{\bf Proposition}}
\newtheorem{statement}   {{\bf Statement}}

\newcommand{\sphere}       {\text{\sc Sphere}\xspace}

\newcommand{\card}[1]{\lvert #1\rvert}

\newcommand{\R}{\mathbb{R}}

\newcommand{\ab}{\hspace{0.125em}}                        

\newcommand{\ie}{\hbox{i.\ab e.}\xspace}                  

\newcommand{\eg}{\hbox{e.\ab g.}\xspace}                  

\title{Finite First Hitting Time\\versus Stochastic Convergence
    in\\Particle Swarm Optimisation\footnote{
This is an extended version of a paper published in MIC 2011 \cite{LehreWitt2011MIC}.
Supported by  
Deutsche Forschungsgemeinschaft (DFG) under grant no. 
WI~3552/1-1.}}

\author{
Per Kristian Lehre
 and Carsten Witt\\
Technical University of Denmark\\
Kgs. Lyngby, Denmark\\
\{pkle,cfw\}@imm.dtu.dk
}

\begin{document}

\maketitle
\begin{abstract}
  We reconsider stochastic convergence analyses of particle swarm
  optimisation, and point out that previously obtained parameter
  conditions are not always sufficient to guarantee mean square
  convergence to a local optimum. We show that stagnation can in fact
  occur for non-trivial configurations in non-optimal parts of the
  search space, even for simple functions like \sphere. The
  convergence properties of the basic PSO may in these situations
  be detrimental to the goal of optimisation, to discover a
  sufficiently good solution within reasonable time.
  To characterise optimisation ability of algorithms, we suggest the
  expected first hitting time (FHT), \ie, the time until a search
  point in the vicinity of the optimum is visited. It is shown that a
  basic PSO may have infinite expected FHT, while an algorithm
  introduced here, the Noisy PSO, has finite expected FHT on some
  functions.
\end{abstract}

\section{Introduction}
Particle Swarm Optimisation (PSO) is an optimisation technique for
functions over continuous spaces introduced by Kennedy and Eberhart
\cite{KennedyEberhartPSO}. The algorithm simulates the motions of a
swarm of particles in the solution space. While limited by inertia,
each particle is subject to two attracting forces, towards the best
position $P_t$ visited by the particle, and towards the best position
$G_t$ visited by any particle in the swarm. The update equations for
the velocity $V_t$ and the position $X_t$ are given in Algorithm
\ref{alg:standard-pso} in Section \ref{sec:preliminaries}.
The inertia
factor $\omega$, and the acceleration coefficients $\varphi_1$ and
$\varphi_2$ are user-specified parameters.  The algorithm only uses
objective function values when updating $G$ and $P$, and does not require any
gradient information. So the PSO is a black-box algorithm \cite{DrosteJansenWegener2006}. It is
straightforward to implement and has been applied successfully in many
optimisation domains.  Despite its popularity, the theoretical
understanding of the PSO remains limited. In particular, how do the
parameter settings influence the swarm dynamics, and in the end, the
performance of the PSO?

One of the best understood aspects of the PSO dynamics are the
conditions under which the swarm stagnates into an equilibrium point.
It is not too difficult to see (e.\,g., \cite{ClercKennedyPSOExplosion}) 
that velocity explosion can only be avoided
when the inertia factor is bounded by
\begin{align}
  |\omega| < 1.\label{eq:conv-cond-omega}
\end{align}
The magnitude of the velocities still depends heavily on how the global
$G_t$ and local best $P_t$ positions evolve with time $t$, which again
is influenced by the function that is optimised. To simplify the
matters, it has generally been assumed that the swarm has entered a
stagnation mode, where the global and local best particle positions
$G_t$ and $P_t$ remain fixed. Under this assumption, there is no
interaction between the particles, or between the problem dimensions,
and the function to be optimised is irrelevant. The swarm can
therefore be understood as a set of independent, one-dimensional
processes.

An additional simplifying assumption made in early convergence
analyses was to disregard the stochastic factors $R$ and $S$,
replacing them by constants
\cite{vandenBerghPhD,ClercKennedyPSOExplosion}. 
Trelea \cite{TreleaPSOconvergence} analysed
the 1-dimensional dynamics under this assumption, showing that
convergence to the equilibrium point 
\begin{align}
P_e= (\varphi_1 P + \varphi_2 G)/(\varphi_1+\varphi_2)\label{eq:equilibrium}
\end{align}
occurs under condition
(\ref{eq:conv-cond-omega}) and
\begin{align}
  0 < \varphi_1 + \varphi_2 & < 4(1+\omega). \label{eq:conv-cond-deterministic}
\end{align}

\begin{figure}
  \centering
  \includegraphics[width=5cm]{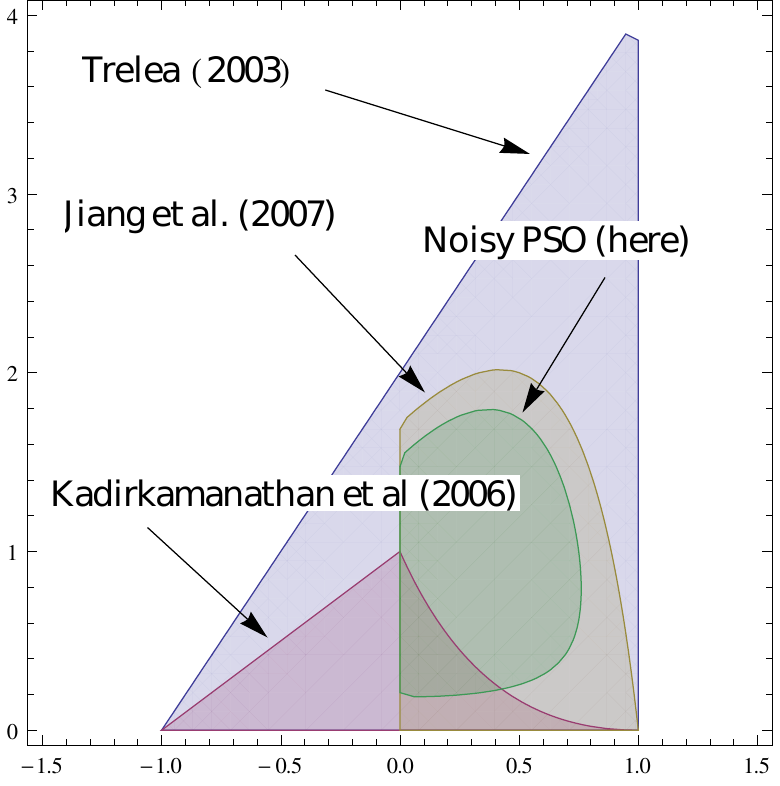}
  \caption{\label{fig:convergence-regions}
    Comparison of convergence regions. Noisy PSO indicates when
    the precondition $f(1)>1/3$ of Theorem~\ref{theo:main-theo-noisy} holds.    
    ($x$-axis: $\omega$, $y$-axis: $\varphi=\varphi_1=\varphi_2$).}
\end{figure}

Kadirkamanathan et al. \cite{Kadirkamanathan2006Stability} were among
the first to take the stochastic effects into account, approaching the
dynamics of the global best particle position (for which $P=G$) from a
control-theoretic angle. In particular, they considered asymptotic
Lyapunov stability of the global best position, still under the
assumption of fixed $P$ and $G$. Informally, this stability condition
is satisfied if the global best particle always converges to the
global best position when started nearby it.  Assuming a global best
position in the origin, their analysis shows that condition
(\ref{eq:conv-cond-omega}), $\omega\neq 0$, and
\begin{align}
  \varphi_1 + \varphi_2 & <2
  (1-2|\omega|+\omega^2)/(1+\omega) \label{eq:conv-cond-lyapunov}
\end{align}
are sufficient to guarantee asymptotic Lyapunov stability of the
origin.
These conditions are not necessary, and are conservative.
Another stochastic mode of convergence considered, is convergence in
mean square (also called second order stability) to a point $x^*$,
defined as
$
  \lim_{t\rightarrow\infty}\expect{|X_t-x^*|^2}=0.
$ 
Mean square convergence to $x^*$ implies that the expectation of the
particle position converges to $x^*$, while its variance converges to
0. It has been claimed that all particles in the PSO converges in mean
square to the global best position if the parameter triplet
$\omega,\varphi_1,\varphi_2$ is set
appropriately. Jiang et al. \cite{JiangLuoYangPSO} derived recurrence equations for
the sequences $\expect{X_t}$ and $\var{X_t}$ assuming fixed $G$ and
$P$, and determined conditions, \ie a convergence region, under which
these sequences are convergent.  The convergence region considered in
\cite{JiangLuoYangPSO} is strictly contained in the convergence region
given by the deterministic condition
(\ref{eq:conv-cond-deterministic}). For positive $\omega$, the
Lyapunov stability region described by condition
(\ref{eq:conv-cond-lyapunov}) is strictly contained in the mean square
stability region. Given the conditions indicated in
Figure~\ref{fig:convergence-regions}, the expectation will converge to
$P_e$ (as in Eq. (\ref{eq:equilibrium})), while the variance will
converge to a value which is proportional to $(G-P)^2$. It is claimed
that the local best $P$ converges to $G$, which would imply that the
variance converges to 0.  However, as we will explain in later
sections, this is not generally correct. We will discuss further
assumptions that are needed to fix the claim of
\cite{JiangLuoYangPSO}.
%
Wakasa et al. \cite{Wakasa2010PSO} pointed out an
alternative technique for determining mean square stability of
specific parameter triplets. They showed that this problem, and other
problems related to the the PSO dynamics, can be reduced to checking
the existence of a matrix satisfying an associated linear matrix
inequality (LMI).  This is a standard approach in control theory, and
is popular because the reduced LMI problem can be solved efficiently
using convex- and quasi-convex optimisation techniques
\cite{Boyd1994LMI}. Wakasa et al. \cite{Wakasa2010PSO} obtained explicit
expressions for the mean square stability region, identical to the
stability region obtained in \cite{JiangLuoYangPSO}, using this
technique.
Assuming stagnation, Poli \cite{Poli2008PSOMoments} provided recurrence
equations for higher moments (\eg skewness and kurtosis) of the
particle distribution.  The equations for the $m$-th moment are
expressed with an exponential number of terms in $m$, but can be
solved using computer algebra systems for not too high moments.

Recently, there has been progress in removing the stagnation
assumption on $P$ and $G$. Building on previous work by
Brandst{\"a}tter and Baumgartner \cite{Brandstatter2002MassSpring}, 
Fern\'andez-Mart\'inez and Garc\'ia-Gonzalo \cite{Martinez2011PSO} interpret
the PSO dynamics as a discrete-time approximation of a certain
spring-mass system. From this mechanical interpretation follows
naturally a generalisation of the PSO with adjustable time step
$\Delta t$, where the special case $\Delta t=1$ corresponds to the
standard PSO. In the limit where $\Delta t\rightarrow 0$, one obtains
a continuous-time PSO governed by stochastic differential
equations. They show that dynamic properties of the discrete-time PSO
approach those of the continuous-time PSO when the time step
approaches 0.

While theoretical research on PSO has mainly focused on convergence,
there may be other theoretical properties that are more relevant in
the context of optimisation.  The primary goal in optimisation is to
obtain a solution of acceptable quality within reasonable time.
Convergence may be neither sufficient, nor necessary to reach this
goal. In particular, convergence is insufficient when stagnation
occurs at non-optimal points in the solution space. Furthermore,
stagnation is not necessary when a solution of acceptable quality has
been found.

As an alternative measure, we suggest to consider for arbitrarily
small $\epsilon>0$ the expected time until the algorithm for the first
time obtains a search point $x$ for which $|f(x)-f(x^*)|<\epsilon$,
where $f(x^*)$ is the function value of an optimal search point, where
time is measured in the number of evaluations of the objective
function. We call this the \emph{expected first hitting time} (FHT)
with respect to $\epsilon$. As a first condition, it is desirable to
have finite expected FHT for any constant $\epsilon>0$. Informally, this
means that the algorithm will eventually find a solution of acceptable
quality. Secondly, it is desirable that the growth of the expected FHT is
upper bounded by a polynomial in $1/\epsilon$ and the number of
dimensions $n$ of the problem. Informally, this means that the
algorithm will not only find a solution of acceptable quality, but
will do so within reasonable time. 

Some work has been done in this direction. Sudholt and Witt
\cite{SudholtWittBinPSO2010} studied the runtime of the Binary PSO, \ie in
a discrete search space. Witt \cite{WittWhyStandard} considered the
Guaranteed Convergence PSO (GCPSO) with one particle on the \sphere
function, showing that if started in unit distance to the optimum,
then after $O(n\log(1/\epsilon))$ iterations, the algorithm has
reached the $\epsilon$-ball around the optimum with overwhelmingly
high probability. The GCPSO avoids stagnation by resetting the global
best particle to a randomly sampled point around the best found
position. The behaviour of the one-particle GCPSO therefore resembles
the behaviour of a (1+1)~ES, and the velocity term does
not come into play. In fact, the analysis has some similarities 
with the analysis by J{\"a}gersk{\"u}pper \cite{Jens2007TCS}.

The objectives of this paper are three-fold. Firstly, in
Section~\ref{sec:stagnation}, we show that the expected first hitting
time of a basic PSO is infinite, even on the simple \sphere
function. Secondly, in Section \ref{sec:conv-analys-jiang}, we point
out situations where the basic PSO does not converge in mean square
to the global best particle (which needs not be a global optimum), 
despite having parameters in the
convergence region. 
We discuss what extra conditions are needed to
ensure mean square convergence. Finally, in
Section~\ref{sec:noisy-pso}, we consider a Noisy PSO which we prove
to have finite expected FHT on the 1-dimensional \sphere function.
Our results also hold for any strictly increasing transformation of
this function because the PSO is a comparison-based algorithm.





\section{Preliminaries}\label{sec:preliminaries}
In the following, we consider minimisation of functions.  A basic PSO
with swarm size $m$ optimising an $n$-dimensional function
$f:\mathbb{R}^n\rightarrow\mathbb{R}$ is defined below. This PSO
definition is well-accepted, and called Standard PSO by
Jiang et al. \cite{JiangLuoYangPSO}. The position and velocity of particle
$i\in[m]$ at time $t\geq 0$ are represented by the pair of vectors
$X_{t}^{(i)} = (X_{t,1}^{(i)},\dots, X_{t,n}^{(i)})$ and $V_{t}^{(i)}
= (V_{t,1}^{(i)},\dots, V_{t,n}^{(i)})$. The parameter $\alpha>0$
bounds the initial positions and velocities.
\begin{algorithm}
  \caption{Basic PSO\label{alg:standard-pso}}
  \begin{algorithmic}
    \FOR{each particle $i\in [m]$, and dimension $j\in[n]$}
    \STATE $X^{(i)}_{0,j},V^{(i)}_{0,j}\sim\unif[-\alpha,\alpha]\;$
           $\; P^{(i)}_{0,j}=X^{(i)}_{0,j}$\\
    \ENDFOR
    \STATE $G_{0}   = \argmin \{ f(P_{0}^{(1)}), \dots, f(P_{0}^{(m)}) \}$\\
    \FOR{$t=0,1,\dots$ until termination condition satisfied}
    \FOR{each particle $i\in [m],$ and dimension $j\in [n]$}
    \STATE\vspace{-0.5cm}
    \begin{align}
      V^{(i)}_{t+1,j}  & = \omega  V^{(i)}_{t,j} \
         + \varphi_1  R^{(i)}_{t,j} \left(P^{(i)}_{t,j}-X^{(i)}_{t,j}\right)
        + \varphi_2 S^{(i)}_{t,j} \left(G_{t,j}-X^{(i)}_{t,j}\right),\label{pso:eq:velocity}\\
       X^{(i)}_{t+1,j}  & = X^{(i)}_{t,j} + V^{(i)}_{t+1,j},\quad
       \text{ where}\quad R^{(i)}_{t,j},S^{(i)}_{t,j} \sim\unif[0,1].
     \end{align}\vspace{-0.5cm}
    \ENDFOR
     \begin{align*}
      P^{(i)}_{t+1}  & = \argmin \{ f(X^{(i)}_{t}), f(P^{(i)}_{t}) \}
      \quad\text{ and }\\
      G_{t+1}   & = \argmin \{ f(P_{t}^{(1)}), \dots, f(P_{t}^{(m)}) \}.       
     \end{align*}
    \ENDFOR
  \end{algorithmic}
\end{algorithm}

Assume that a function $f:\mathbb{R}^n\rightarrow\mathbb{R}$ has at
least one global minimum $x^*$. Then for a given $\epsilon>0$, the
\emph{first hitting time} (FHT) of the PSO on function $f$ is defined
as the number of times the function $f$ is evaluated until the swarm
for the first time contains a particle $x$ for which
$|f(x)-f(x^*)|<\epsilon$. We assume that the PSO is implemented such
that the function $f$ is evaluated no more than $m$ times per time
step $t$.
As an example function, we consider the \sphere problem, which for all
$x\in\mathbb{R}^n$ is defined as
$\sphere(x) := \| x\|^2$,
where $\|\cdot\|$ denotes the Euclidian norm. This is a well-accepted benchmark 
problem in convergence analyses and frequently serves as a starting point 
for theoretical analyses.

\section{Stagnation}\label{sec:stagnation}
Particle convergence does not necessarily occur in local optima.
There are well-known configurations, \eg with zero velocities,
which lead to stagnation \cite{vandenBerghPhD}. However, it is not
obvious for which initial configurations and parameter settings the
basic PSO will stagnate outside local optima. Here, it is shown
that stagnation occurs already with 1 dimension for a broad range of
initial parameters. It follows that the expected first hitting time
of the basic PSO can be infinite.

As a first example of stagnation, we consider the basic PSO with swarm
size one on the \sphere problem. Note that it is helpful to first
study the PSO with swarm size one before analysing the behaviour of
the PSO with larger swarm sizes. This is similar to the theory of
evolutionary algorithms (EAs), where it is common to initiate runtime
analyses on the simple (1+1) EA with population size one, before
proceeding to more complex EAs.
 

\begin{proposition}
 \label{prop:standard-oneparticle-infinite-fht}
  The basic PSO with inertia factor $\omega<1$ and one particle ($m=1$)
  has infinite expected FHT on \sphere ($n=1$).
\end{proposition}
\begin{proof}
  We say that the \emph{bad initialisation event} has occurred if the
  initial position and velocity satisfy $X_0>\varepsilon\alpha$ and
  $
    (\varepsilon\alpha-X_0)/(1-\omega) < V_0 < 0.
  $
  This event occurs with positive probability. We claim that if the
  event occurs, then in any iteration $t> 0$
  $
    V_{t}  = V_0\omega^{t-1}
  $,
    and
  $
    X_t    = X_0 + V_0\sum_{i=0}^{t-1} \omega^i.
  $
  If the claim holds, then for all $t\geq 0$, it holds that
  $X_t<X_{t-1}$ and $G_t=X_t$. Therefore,
  \begin{align*}
    G_t & > X_0 + V_0\sum_{i=1}^\infty \omega^t
            > X_0 + \varepsilon\alpha-X_0 = \varepsilon\alpha.
  \end{align*}
  and the proposition follows.
  Note that since $G_t=X_t$ 
  for each $t>0$, the velocity reduces to $V_t=\omega V_{t-1}$.

  The claim is proved by induction on $t$. The base case $t=2$ clearly
  holds, because $V_1 = V_0\omega$ and $X_1 = X_0 + V_0\omega$.
  Assume the claim holds for all iterations smaller than $t$.  By
  induction, it holds that
  $ V_t=\omega V_{t-1} = \omega^{t-1} V_0.$
  Therefore, by the induction hypothesis,
  \begin{align*}
    X_t  = X_{t-1} + V_{t} 
         = X_0 + V_0\sum_{i=1}^{t-2}\omega^i + V_0\omega^{t-1}
         = X_0 +  V_0\sum_{i=1}^{t-1}\omega^i.
  \end{align*}
  The claim now holds for all $t>0$.
  The expected FHT conditional on the bad initialisation event is
  therefore infinite. However, the bad initialisation event occurs
  with positive probability, so the unconditional expected FHT is 
  infinite by the law of total probability.
\end{proof}

We prove that the stagnation on \sphere illustrated 
in Proposition~\ref{prop:standard-oneparticle-infinite-fht} is not an artefact of 
a trivial swarm size of~$1$. In the following theorem, we prove
stagnation for a swarm of size~$2$ and think that the ideas 
can be generalised to bigger swarm sizes. We allow any initialisation of 
the two particles that are sufficiently far away from the optimum. It is 
assumed that both velocities are non-positive  in the initialisation step, 
which event occurs with constant 
probability for uniformly drawn velocities.

\begin{theorem}
\label{theo:pso-2-particle-divergence}
Consider the basic PSO with two particles on the one-dimensional 
\sphere.   
If $\omega < 1$, $1<\varphi_2 < 2$, $V_0^{(1)}, V_0^{(2)}\le 0$,  
$\kappa<1$ where
\begin{align*}
 \kappa := \frac{\varphi_2^2-2\varphi_2+2+2\omega\varphi_2}{4\varphi_2}
          +\frac{\sqrt{(\varphi_2^2-2\varphi_2+2\omega\varphi_2+2)(\varphi_2^2+6\varphi_2+2\omega\varphi_2+2)}}{4\varphi_2},
\\\text{and}\quad
X_0^{(1)},X_0^{(2)}    
 > 2\epsilon + 2\varphi_2\cdot\left(\frac{\card{X_0^{(2)} - X_0^{(1)}}+\card{V_0^{(1)}}+\card{V_0^{(2)}}}{(1-\omega)(1-\kappa)}\right)   
\end{align*}
all hold together,
then the expected FHT for the $\epsilon$-ball around the optimum is infinite.
\end{theorem}

The conditions are fulfilled, \eg, if $\varphi_2=1.5$, $\omega = 0.07$, $\epsilon=0.5$, $V_{0}^{(1)}=V_0^{(2)}=-1$, $X_{0}^{(1)}=184$, and 
$X_{0}^{(2)}=185$.
For a proof, we note that the assumed initialisation with positive
particle positions, negative velocities and $\alpha$ sufficiently large
makes the sequences $X_t^{(i)}$, $i=1,2$, non-increasing provided no negative values are reached. Furthermore, the update 
equation for the velocities will then consist of three random non-positive terms, which means that 
velocities remain negative.
In Lemma \ref{lem:bound-edt}, we focus on the distance 
$D_t:=X_t^{(2)} - X_t^{(1)}$ of the particles and show that its expectation converges absolutely to
zero. The proof of this lemma makes use of Lemma \ref{lem:fibonacci}, which gives a
closed-form solution to a generalisation of the Fibonacci-sequence.
In another lemma, 
we consider the absolute velocities over time and show that the series 
formed by these also converges in expectation. The proof of the theorem 
will be completed by applications of Markov's inequality.

\begin{lemma}
\label{lem:fibonacci}
  For any real $c>0$, there exists two reals $A$ and $B$ such that the
  difference equation $a_n  = c(a_{n-1} + a_{n-2}), n\geq 1,$
  has the solution $a_n = \alpha^{n} A + \beta^{n}B,$
  where
  \begin{align*}
    \alpha  = \frac{c-\sqrt{c(4+c)}}{2},\quad \text{and}\quad
    \beta   =  \frac{c+\sqrt{c(4+c)}}{2}.
  \end{align*}
\end{lemma}
\begin{proof}
  The proof is by induction over $n$. The lemma can always be
  satisfied for $n=1$ and $n=2$ by choosing appropriate $A$
  and $B$. Hence, assume that the lemma holds for all $i<n$ for some
  $A$ and $B$. Note that 
  \begin{align*}
    c(\alpha+1) & = \frac{c^2-c\sqrt{c(4+c)}+2c}{2} = \alpha^2,\quad\text{ and}\\
    c(\beta+1) & = \frac{c^2+c\sqrt{c(4+c)}+2c}{2} = \beta^2
  \end{align*}
  It therefore follows by the induction hypothesis that
  \begin{align*}
    a_n & = c(a_{n-1}+a_{n-2})\\
    & = c(\alpha^{n-1}A + \beta^{n-1}B +
    \alpha^{n-2}A + \beta^{n-2}B)\\
    & = c\alpha^{n-2}(\alpha+1)A + 
    c\beta^{n-2}(\beta+1)B\\
    & = \alpha^{n} A + \beta^{n}B.
\end{align*}
\end{proof}

\begin{lemma}
\label{lem:bound-edt}
Given $t\ge 1$, suppose that for all $s\le t$ it holds that 
$X_s^{(1)}, X_s^{(2)}\ge 0$ and $V^{(1)}_s,V^{(2)}_s\le 0$. 
Then 
$E(\card{D_t}) \le \kappa^{t} (2\card{D_0}+ V_0^{1}-V_0^{2})$.  
\end{lemma}
\begin{proof}
The proof is mainly based on an inspection of the update equation of PSO. 
 The aim is to obtain a recurrence 
for $E(\card{D_t})$, where we have to distinguish between two cases. We abbreviate $\varphi=\varphi_2$ and $S=S^{(\cdot)}_t$ in the following.

If $X_t^{(1)}\le X_t^{(2)}$, then $G_t=X_t^{(1)}$ and the update equations are
\begin{align*}
V^{(1)}_{t+1} & = \omega V^{(1)}_{t}\\
V^{(2)}_{t+1} & = \omega V^{(2)}_{t}+S\varphi (X_t^{(1)} - X_t^{(2)}) = 
\omega V^{(2)}_{t}-S\varphi D_t,
\end{align*}
which means
$
D_{t+1} = X_{t+1}^{(2)} - X_{t+1}^{(1)} = X_{t}^{(2)} - X_{t}^{(1)}  
-S\varphi D_t + \omega(V_{t}^{(2)} - V_{t}^{(1)})  .
$
Since
$
V_{t+1}^{(i)} = X_{t+1}^{(i)} - X_{t}^{(i)},
$
for $i=1,2$, 
we obtain
$
V_t^{(2)}-V_t^{(1)} = D_t-D_{t-1},
$
for $t\ge 0$, where we define $D_{-1}=D_0-V_0^{(2)}+V_0^{(1)}$ 
to make the equation apply also for $t=0$.
Together, 
this gives us
\begin{equation}
\label{eq:difference-dt}
D_{t+1} = D_{t} - S\varphi D_t + \omega(D_t - D_{t-1}),
\end{equation}
If $X_t^{(1)}\ge X_t^{(2)}$, then 
the update equations are  
$
V^{(1)}_{t+1}  = \omega V^{(1)}_{t} +S\varphi (X_t^{(2)} - X_t^{(1)})
 = 
\omega V^{(1)}_{t}+S\varphi D_t
$
and
$V^{(2)}_{t+1}  = \omega V^{(2)}_{t}$,
which again results in~\eqref{eq:difference-dt} and 
finishes the case analysis.
Taking absolute values on both sides of~\eqref{eq:difference-dt} and 
applying the triangle inequality to the right-hand side, we get
\[
\card{D_{t+1}} \le \card{(1-S\varphi+\omega)} \card{D_t}  + \card{\omega}\card{ D_{t-1}}.
\]
After taking the expectation and noting that 
$\omega>0$, we 
have
\[
E(\card{D_{t+1}}\mid D_{t},D_{t-1}) \le (E(\card{1-S\varphi})+\omega) \card{D_t} + \omega \card{D_{t-1}},
\]
which implies 
$
E(\card{D_{t+1}}) \le (E(\card{1-S\varphi})+\omega) E(\card{D_t}) + \omega E(\card{D_{t-1}}),
$
as the right-hand side is linear in both $\card{D_{t-1}}$ and $\card{D_{t}}$.
We are left with an estimate for $E(\card{1-S\varphi})$. By the law 
of total probability, 
\begin{align*}
E(\card{1-S\varphi})  & = 
   E(1-S\varphi \mid 1-S\varphi \ge 0)\Pr(1-S\varphi\ge 0) \\ 
&\quad +  E(S\varphi-1 \mid 1-S\varphi \le 0)\Pr(1-S\varphi\le 0)
\end{align*}
Since $S$ is uniformly distributed and $\varphi>1$, the first conditional expectation 
is $1/2$, while the second conditional expectation is $(\varphi-1)/2$. The probabilities 
for the conditions to occur are $1/\varphi$ and $1-1/\varphi$, respectively. This 
results in
\[
E(\card{1-S\varphi}) = \frac{1}{2\varphi} + \left(1-\frac{1}{\varphi}\right)\cdot \frac{\varphi-1}{2}
= \frac{\varphi^2-2\varphi+2}{2\varphi},
\]
which finally gives us the following recurrence on $E(\card{D_t})$:
\[
E(\card{D_{t+1}}) \le \left(\frac{\varphi^2-2\varphi+2}{2\varphi} + \omega\right) (E(\card{D_{t}}) + E(\card{D_{t-1}})).
\]
Introducing $D^*_t:=E(\card{D_t})$ and using $\lambda=\frac{\varphi^2-2\varphi+2}{\phi}+2\omega$, we
have in more compact form that
\[
D^*_{t+1} \le \frac{\lambda}{2} (D^*_t + D^*_{t-1})
\]
for $t\ge 0$.
Solving this recursion (noting that all terms are positive) using 
Lemma~\ref{lem:fibonacci} yields for $t\ge 1$ that
\[
D^*_{t} \le \left(\frac{\lambda-\sqrt{8\lambda+\lambda^2}}{4}\right)^t D^*_{-1} + \left(\frac{\lambda+\sqrt{8\lambda+\lambda^2}}{4}\right)^t D^*_0.
\]

Note that $\kappa=\frac{\lambda+\sqrt{8\lambda+\lambda^2}}{4} < 1$ if and only if $\lambda<1$. Furthermore, 
the factor in front of $D^*_{-1}$ has clearly smaller absolute value than $\kappa$. We obtain
$
D^*_{t} \le \kappa^t (D^*_{-1} + D^*_0)  \le \kappa^t (2D^*_0 + V_0^{1}-V_0^{2})
$
which we wanted to show.
\end{proof}

The following lemma uses the previous bound on $E(\card{D_t})$ to show 
that the expected sum of velocities converges absolutely over time. This means 
that the maximum achievable progress is bounded in expectation. As an example, when 
choosing $\varphi_2=1.5$ and $\omega=0.07$ in the following, 
we obtain a value of about $191 (\card{D_{0}}+\card{V_0^{(1)}}+\card{V_0^{(2)}})$ for this bound.

\begin{lemma}
\label{lem:bound-sumevt}
Suppose the prerequisites of Lemma~\ref{lem:bound-edt} apply. Then 
for $i=1,2$ it holds that 
\begin{align*}
\sum_{t=0}^\infty E(\card{V_t^{(i)}}) \le \left(\frac{2\varphi_2}{(1-\omega)(1-\kappa)}\right) (\card{D_{0}}+\card{V_0^{(1)}}+\card{V_0^{(2)}}).  
\end{align*}
\end{lemma}
\begin{proof}
For notational convenience, we drop the upper index and 
implicitly show the following for both $i=1$ and $i=2$. 
According to the update equation of PSO, we have
$\card{V_{t+1}} \le w\card{V_t} + \varphi\card{D_{t}}$ for $t\ge 0$, using 
$\varphi:=\varphi_2$.
Resolving the recurrence yields for $t\ge 1$ that 
\[
\card{V_t} 
\le \omega^{t}\card{V_0} + \varphi \sum_{s=0}^{t-1} \omega^{s} \card{D_{t-1-s}} 
= \omega^{t}\card{V_0} + \varphi\sum_{s=0}^{t-1} \omega^{t-1-s} \card{D_s}.
\]
Hence,
\begin{align*}
\sum_{t=1}^{\infty}\card{V_t} & \le \sum_{t=1}^{\infty} \left(\omega^{t}\card{V_0} + \varphi\sum_{s=0}^{t-1} \omega^{t-1-s} \card{D_s}\right)\\
& \le \frac{\card{V_0}}{1-\omega} + \varphi\left(\sum_{t=0}^\infty \left(\sum_{s\ge t} \omega^s\right) \card{D_t}\right) 
 \le \frac{\varphi}{1-\omega}\left(\card{V_0} + \sum_{t=0}^\infty \card{D_t}\right),
\end{align*}
since $0<\omega<1$ and $\varphi> 1$.
Using the linearity of expectation, 
\[
\sum_{t=0}^\infty E(\card{V_t} )
\le \frac{\varphi}{1-\omega}\left(\card{V_0}+\sum_{t=0}^\infty E(\card{D_t})\right).
\]

By 
Lemma~\ref{lem:bound-edt},
$
E(\card{D_t}) \le   \kappa^{t} (2\card{D_{0}}+V_0^{(1)}-V_0^{(2)}).
$
 Hence, the series 
over the $E(\card{D_t})$ 
converges according to
\[
\sum_{t=0}^\infty E(\card{D_t} ) \le \frac{1}{1-\kappa} (2\card{D_{0}}+V_0^{(1)}-V_0^{(2)}),
\]
which yields
\begin{align*}
 \sum_{t=0}^\infty E(\card{V_t} ) & \le 
\frac{\varphi}{1-\omega} (V_0^{(1)} + V_0^{(2)}) + 
\frac{\varphi}{1-\omega} \frac{1}{1-\kappa} (2\card{D_{0}}+V_0^{(1)}-V_0^{(2)})\\
& \le 
\left(\frac{\varphi}{(1-\omega)(1-\kappa)}\right) \cdot (2\card{D_{0}} + 2\card{V_0^{(1)}}+2\card{V_0^{(2)}}),
\end{align*}
where we have used $\kappa<1$.
\end{proof}

We are ready to prove Theorem~\ref{theo:pso-2-particle-divergence}.
\begin{proof}[Proof of Theorem~\ref{theo:pso-2-particle-divergence}]
Throughout this proof, we suppose  the prerequisites from Lemma~\ref{lem:bound-edt} 
to hold, which, as we will show, is true for an infinite number of steps with 
constant probability. 

For any finite~$t$, Lemma~\ref{lem:bound-sumevt} and linearity 
of expectation 
yield for $i=1,2$ that 
\[
\mathord{E}\mathord{\left(\sum_{s=0}^t \card{V_s^{(i)}} \right)}
 \le \left(\frac{2\varphi_2}{(1-\omega)(1-\kappa)}\right) \cdot (\card{D_{0}}+\card{V_0^{(1)}}+\card{V_0^{(2)}}),
 \]
which by Markov's inequality means that the event 
\[
\sum_{s=0}^{t} \card{V_s^{(i)}}  \le  \epsilon + \left(\frac{2\varphi_2}{(1-\omega)(1-\kappa)}\right) \cdot (\card{D_{0}}+\card{V_0^{(1)}}+\card{V_0^{(2)}})
\]
occurs with a positive probability that does not depend on~$t$. Given the assumed initial values of $X_{0}^{(i)}$, the 
$\epsilon$\nobreakdash-ball around the optimum is not reached if the event occurs. Hence, there is a minimum probability 
$p^*$ such that for any finite 
number of steps~$t$, the probability of not hitting the $\epsilon$\nobreakdash-ball within $t$~steps 
is at least~$p^*$. Consequently, 
the expected first hitting time of the $\epsilon$\nobreakdash-ball is infinite.
\end{proof}

\section{Mean Square Convergence}
\label{sec:conv-analys-jiang}
As mentioned in the introduction, there exist several convergence
analyses using different techniques that take into account the
stochastic effects of the algorithm. The analysis by
Jiang et al. \cite{JiangLuoYangPSO} is perhaps the one where the proof of mean
square convergence follows most directly from the definition.  They
consider the basic PSO and prove the following statement (Theorem~5
in their paper):

\begin{statement}
\label{statement:jiang-theorem}
Given $\omega,\varphi_1,\varphi_2\ge 0$, if $0\le \omega<1$, 
$\varphi_1+\varphi_2>0$, and $0<-(\varphi_1+\varphi_2)\omega^2 + 
\left(\frac{1}{6}\varphi_1^2+\frac{1}{6}\varphi_2^2+\frac{1}{2}\varphi_1\varphi_2\right)\omega
+ \varphi_1+\varphi_2 - \frac{1}{3}\varphi_1^2 - \frac{1}{3}\varphi_2^2 - \frac{1}{2}\varphi_1\varphi_2 
< \frac{\varphi_2^2(1+\omega)}{6}$ are all satisfied,
the basic particle swarm 
system determined by parameter tuple $\{\omega,c_1,c_2\}$ will converge in mean square to 
$G$.
\end{statement}

This statement is claimed to hold for any fitness function and for any initial 
swarm configuration. However, as acknowledged by the corresponding author \cite{JiangPersComm}, 
there is an error in the proof of the above statement, which is actually wrong 
without additional assumptions.

Intuitively Statement~\ref{statement:jiang-theorem} makes sense for well-behaved, 
continuous functions like \sphere. However, in retrospect, 
it is not too difficult to set up 
artificial fitness functions and swarm configurations where the statement 
is wrong: Let us consider the one-dimensional function $f: \R\to\R$ defined by
$f(0)=0$, $f(1)=1$, and $f(x)=2$ for all $x\in\R\setminus\{0,1\}$,
which is to be minimised.

Assume a swarm of two particles, where the first one has position~$0$,
which is then its local best and the global best. Furthermore, assume
velocity 0 for this particle, i.e., it has stagnated. Formally, 
$X_0^{(1)}=P_0^{(1)}=G_0=V_0^{(1)}=0$.
Now let us say the second particle has current and local best position
1 and 
velocity 0, formally $X_0^{(2)}=P_0^{(2)}=1$ and $V_0^{(2)}=0$.
This particle will now be attracted by a weighted combination 
of local and global best, e.\,g.\ the point $0.5$ if both learning rates are 
the same. The problem is that the particle's local best almost surely will 
never be updated again since the probability of sampling either local 
best or global best is $0$ if the sampling distribution is uniform on an 
interval of positive volume or is the sum of
two such distributions, as it is defined in the basic PSO. The sampling distribution might be deterministic 
because both $P_t^{(i)}-X_t^{(i)}$ and $G_t-X_t^{(i)}$ might be~$0$, 
but then then the progress corresponds to the 
last velocity value, which again was either obtained 
according to the 
sum of two uniform distributions or was already~$0$.
The error in the analysis is hidden in the proof of Theorem~4 
in \cite{JiangLuoYangPSO}, where $\Pr(X_t=G)>0$ is concluded 
even though $G$ might be in a null set. Nevertheless, important 
parts of the preceding analysis can be saved and a theorem 
on convergence can be proved under additional assumptions 
on the fitness function. In the following, we describe the 
main steps in the convergence analysis by Jiang et al. \cite{JiangLuoYangPSO}. 
A key idea in \cite{JiangLuoYangPSO} is to consider a one-dimensional algorithm 
and an arbitrary particle, assuming that the local best for this particle and global best 
do not change. Then a 
recurrence 
relation is obtained as follows:
$
X_{t+1} = (1+\omega-(\varphi_1 R_t + \varphi_2 S_t)) X_t 
- \omega X_{t-1} + \varphi_1 R_t P + \varphi_2 S_t G ,
$
where we dropped the index denoting the arbitrary particle we have chosen, and 
the time index for local and global best. The authors proceed by deriving sufficient conditions for the 
sequence of expectations $E(X_t)$, $t\ge 1$, to converge (Theorem~$1$ in their paper).

\begin{lemma}
Given $\omega, \varphi_1,\varphi_2\ge 0$, if and only if $0\le \omega < 1$ 
and $0<c_1+c_2 < 4(1+\omega)$, the iterative process $E(X_t)$ is guaranteed 
to converge to $(\varphi_1 P + \varphi_2 G)/(\varphi_1+\varphi_2)$.
\end{lemma} 

Even though a process converges in expectation, its variance might diverge, 
which intuitively means that it becomes more and more unlikely to observe
the actual process in the vicinity of the expected value. 
Another major achievement by Jiang et al. \cite{JiangLuoYangPSO} is to
study the variances $\Var(X_t)$ of the still one-dimensional 
process. By a clever analysis of a recurrence of order~$3$, they 
obtain the following lemma (Theorem~$3$ in their paper).

\begin{lemma}
\label{lem:jiang-theorem3}
Given $\omega,\varphi_1,\varphi_2\ge 0$, if and only if
$0\le \omega<1$, $\varphi_1 + \varphi_2\ge 0$ and $f(1)>0$ 
are all satisfied together, iterative process 
$\Var(X_t)$ is guaranteed to converge to $\frac{1}{6}(\varphi_1\varphi_2/(\varphi_1+\varphi_2))^2 + 
(G-P)^2 (1+\omega) /f(1)$, 
where
$f(1)= 
-(\varphi_1+\varphi_2)\omega^2 + 
(\frac{1}{6}\varphi_2^2 + \frac{1}{2} \varphi_1^2\varphi_2^2) \omega
+ \varphi_1 + \varphi_2 - \frac{1}{3} \varphi_1^2 - \frac{1}{3} \varphi_2^2 
-\frac{1}{2}\varphi_1\varphi_2$.
\end{lemma}

Lemma~\ref{lem:jiang-theorem3} means that the variance is proportional to 
$(P-G)^2$. However, in contrast to what Jiang et al. \cite{JiangLuoYangPSO} 
would like to achieve in their Theorem~4, we do not see how to 
prove that the variance approaches~$0$ for every particle. Clearly, 
this 
happens for the global best particle under the assumption that 
no further improvements of the global best are found. We do not 
follow this approach further since we are interested in PSO 
variants that converge to a local optimum.

\section{Noisy PSO}
\label{sec:noisy-pso}

The purpose of this section is to consider a variant of the basic PSO
that includes a noise term. This PSO, which we call the Noisy PSO, is
defined as in Algorithm~\ref{alg:standard-pso}, except that
Eq. (\ref{pso:eq:velocity}) is replaced by the velocity equation
$
  V^{(i)}_{t+1,j}   =   \omega V^{(i)}_{t,j} 
                     + \varphi_1 R^{(i)}_{t,j} (P^{(i)}_{t,j}-X^{(i)}_{t,j})  
                     + \varphi_2 S^{(i)}_{t,j} (G_{t,j}-X^{(i)}_{t,j})
                     + \Delta_{t,j}^{(i)},
$ 
where the extra noise term $\Delta_{t,j}^{(i)}$ has uniform
distribution on the interval $[-\delta/2,\delta/2]$.  Note that our
analysis seems to apply also when the uniform distribution is
replaced by a Gaussian one with the same expectation and variance.
The constant parameter $\delta>0$ controls the noise level in the
algorithm.  Due to the uniformly distributed noise term, it is
immediate that the variance of each particle is always at least
$\delta^2/12$. Therefore, the Noisy PSO does not enjoy the mean square
convergence property of the basic PSO. In return, the Noisy PSO does
not suffer the stagnation problems discussed in
Section~\ref{prop:standard-oneparticle-infinite-fht}, and finite
expected first hitting times can in some cases be guaranteed.  The
noisy PSO uses similar measures to avoid stagnation as the GCPSO
mentioned in the introduction. However, our approach is simpler and
treats all particles in the same way. On the other hand, the GCPSO
relies on a specific update scheme for the global best particle.

Our main result considers the simplified case of a one-dimensional 
function 
but takes into account the whole particle swarm. For simplicity, we 
only consider the half-open positive interval by defining
$\sphere^+(x):= \sphere(x)$ if $x\ge 0$, and $\sphere^+(x):=\infty$ otherwise,
which has to be minimised, and assume that at least one particle 
is initialised in the positive region. This event happens with
positive probability for a standardised initialisation scheme.
It seems that our analyses can be adapted to 
the standard $\sphere$ (and order-preserving transformations thereof), 
but changes of sign complicate the analysis 
considerably. Note that the analyses of stagnation in Section~\ref{sec:stagnation} 
only consider positive particle positions and thus apply to 
$\sphere^+$ as well.

\begin{theorem}
\label{theo:main-theo-noisy}
Consider the Noisy PSO on the $\sphere^+$ function 
and assume $G_0\ge 0$. 
If $\delta\le\epsilon, f(1)>1/3,$ and the assumptions from 
Theorems~\ref{thm:exp-xt} and~\ref{thm:var-xt} below hold, 
then the expected first hitting time 
for the interval $[0,\epsilon]$ is finite.
\end{theorem}

The proof of this theorem relies heavily on the convergence 
analysis by Jiang et al. \cite{JiangLuoYangPSO}. We will adapt their 
results to the Noisy PSO. Recall that the only difference 
between the two algorithms is the addition of $\Delta^{(i)}$ 
in the update equation for the particle position. It is important 
to note that $\Delta^{(i)}$ is drawn from $[-\delta/2,\delta/2]$ (considering 
one dimension) fully independently for every particle and time step.
%
As mentioned above, Jiang et al. \cite{JiangLuoYangPSO} consider a one-dimensional algorithm 
and an arbitrary particle, assuming that the local best for this particle and global best 
do not change. Then a 
recurrence 
relation is obtained by manipulating the update equations. Taking this 
approach for the Noisy PSO yields:
$
X_{t+1} \;=\; 
 (1+\omega-(\varphi_1 R_t + \varphi_2 S_t)) X_t 
- \omega X_{t-1} + \varphi_1 R_t P + \varphi_2 S_t G + \Delta_t,
$
where we dropped the index for the dimension, the 
index denoting the arbitrary particle we have chosen and 
the time index for local and global best. This is the same recurrence relation 
as in \cite{JiangLuoYangPSO} except for the addition of $\Delta_t$.
The authors proceed by deriving sufficient conditions for the 
sequence of expectations $E(X_t)$, $t\ge 1$, to converge. Since $E(\Delta_t)=0$, 
the recurrence relation for the expectations is \emph{exactly} the same as with the 
basic PSO and the following theorem can be taken over.

\begin{theorem}\label{thm:exp-xt}
Given $\omega, \varphi_1,\varphi_2\ge 0$, if and only if $0\le \omega < 1$ 
and $0<\varphi_1+\varphi_2 < 4(1+\omega)$, the iterative process $E(X_t)$ is guaranteed 
to converge to $(\varphi_1 P + \varphi_2 G)/(\varphi_1+\varphi_2)$.
\end{theorem} 

The next step is to study the variances $\Var(X_t)$ of the one-dimensional 
process. Obviously, modifications of the original analysis in \cite{JiangLuoYangPSO}
become necessary here. To account for the addition of $\Delta_t$, we replace
Eq.~(11) in the paper\footnote{When referring to the analysis by Jiang
  et al. \cite{JiangLuoYangPSO}, $R_t$ does 
not mean the random factor in the cognitive component, but should 
be understood as defined in their paper.} by
$
Y_{t+1} \;=\; (\psi - R_t) Y_t - \omega Y_{t-1} + Q_t',
$
where $Q_t':=Q_t + \Delta_t$ and $Q_t$ is the original $Q_t$ 
from the paper. Regarding the quantities involving $Q_t'$ in the 
following, we observe that 
$
E(Q_t') = E(Q_t) = 0
$
and 
$\Var(Q_t') 
  = E((Q_t')^2) 
  = E( (Q_t+\Delta_t) (Q_t+\Delta_t)) 
  = E(Q_t^2 + 2\Delta_t Q_t + Q_t^2)
  = E(Q_t^2) + E(\Delta_t^2),
$ 
where we used that $\Delta_t$ is drawn independently of other 
random variables. Finally, we get 
$
E(R_tQ_t') = E(R_t (Q_t+\Delta_t))) = E(R_tQ_t),
$
which means that all following calculations in Section~3.2 
in \cite{JiangLuoYangPSO} may use the same values 
for the variables $R$  and $T$ as before. Only the variable 
$Q$ increases by $E(\Delta^2_t)$. Recall that $\Delta_t \sim U[-\delta/2,\delta/2]$ 
for  constant~$\delta>0$. We obtain $E(\Delta^2_t)=\delta^2/12$. 
Now the iteration 
equation~(17) for $\Var(X_t)$ can be taken over with $Q$ increased by $\delta^2/12$. 
The characteristic equation~$(18)$ remains unchanged and Theorem~$2$ applies 
in the same way as before. Theorem~3 in \cite{JiangLuoYangPSO} is updated 
in the following way and proved as before, except for plugging in the updated 
value of~$Q$.

\begin{theorem}\label{thm:var-xt}
Given $\omega,\varphi_1,\varphi_2\ge 0$, if and only if
$0\le \omega<1$, $\varphi_1 + \varphi_2\ge 0$ and $f(1)>0$ 
are all satisfied together, iterative process 
$\Var(X_t)$ is guaranteed to converge to $(\frac{1}{6}(\varphi_1\varphi_2/(\varphi_1+\varphi_2))^2  
(G-P)^2 (1+\omega)) + \delta^2/12)/f(1)$, 
where
\begin{align*}
f(1)  = 
-(\varphi_1+\varphi_2)\omega^2 + 
\left(\frac{1}{6}\varphi_2^2 + \frac{1}{2} \varphi_1^2\varphi_2^2\right) \omega
 + \varphi_1 + \varphi_2 - \frac{1}{3} \varphi_1^2 - \frac{1}{3} \varphi_2^2 
-\frac{1}{2}\varphi_1\varphi_2.  
\end{align*}
\end{theorem}

As a consequence from the preceding lemma, the variance remains 
positive even for the particle~$i$ that satisfies $P^{(i)}=G$. Under 
simplifying assumptions, we show that this particle allows the 
system to approach the optimum. Later, we will show how to drop 
the assumption.

\begin{lemma}
\label{lem:improvement-deltatenth}
  Assume
  that
  $f(1)>1/3$, that the global and local bests are never updated, and
  that the conditions in Theorem \ref{thm:exp-xt} and Theorem
  \ref{thm:var-xt} hold. Then for all sufficiently small 
  $\varepsilon'>0,$ there exists a
  $t_0>0$ such that
  \begin{align*}
    \forall t\geq t_0\quad \prob{G-\delta \leq X_t\leq
      G-\delta/100+\varepsilon'}\geq 3/100000,
  \end{align*}
  where $X_t\in \R$ is the position of the particle in 
  iteration $t$ for which the local best position equals the 
  global best position.
\end{lemma}
\begin{proof}
  We assume that $G>\varepsilon$, otherwise there is nothing to show.
  Furthermore, $G-\delta$ cannot be negative since
  $\delta\le\varepsilon$.
  We decompose the process by defining
  $Y_t=X_t-\Delta_t$. Our goal is to prove that it is unlikely that
  $Y_t$ is much larger than $G$ using Chebyshev's inequality. We
  therefore need to estimate the expectation and the variance of
  $Y_t$.
  From Theorem~\ref{thm:exp-xt} and the fact that $P=G$ holds for the
  best particle, 
  \begin{align}
    \tlim \expect{Y_t} = \tlim \expect{X_t} - \expect{\Delta_t} = G.  \label{eq:1}
  \end{align}
  To estimate the variance of $Y_t$, first recall that by
  Theorem~\ref{thm:var-xt}, it holds that
  \begin{align}
    \tlim \var{X_t} = \delta^2/12f(1).\label{eq:3}
  \end{align}
Due to the independence of the random variables $Y_t$ and $\Delta_t$, we have
$\var{X_t}=\var{Y_t}+\var{\Delta_t}$. The random variable $\Delta_t$ has 
variance $\delta^2/12$. The limit in (\ref{eq:3}) therefore implies that 
\begin{align}
  \tlim \var{Y_t} 
  & = \tlim\var{X_t} - \var{\Delta_t}
    = \sigma_Y^2, \label{eq:2}
\end{align}
where we have defined
$
  \sigma_Y^2 := \delta^2(1-f(1))/(12f(1)) \leq \delta^2/6.  
$
Combining Eq. (\ref{eq:1}) and Eq. (\ref{eq:2}), yields
$
  \tlim \expect{Y_t}+(6/5)\sqrt{\var{Y_t}} = G + (6/5)\sigma_Y.
$
This limit implies that for any $\varepsilon'>0$, there exists a
$t_0>0$ such that 
$
  \forall t\geq t_0\quad \expect{Y_t}+(6/5)\sqrt{\var{Y_t}} 
   \leq G+(6/5)\sigma_Y+\varepsilon'
   \leq G+0.4899\delta+\varepsilon',
$ 
and analogously $\expect{Y_t}-(6/5)\sqrt{\var{Y_t}} \ge 
G+0.4899\delta+\varepsilon'$.
By the inequality above, and by Chebyshev's inequality, it holds that 
\begin{align*}
  p  := \prob{ |Y_t-G|\geq 0.4899\delta+\varepsilon'}
    \leq \prob{|Y_t-\expect{Y_t}|\geq (6/5)\sqrt{\var{Y_t}}} 
    \leq 25/36.
\end{align*}
Obviously, the larger $Y_t$ is the more restrictive the requirements 
on the outcome of~$\Delta_t$ are. 
Hence, choosing $t$ so large that $\varepsilon'\le (1-0.4899)\delta$ holds, 
we get the desired result
\begin{align*}
  \prob{G-\delta \le X_t<G-\delta/100+\varepsilon'} 
  & \geq \prob{\Delta_t<-\delta(1/100+0.4899)}\cdot(1-p)\\
  & \geq 3/100000.\qedhere
\end{align*}
\end{proof}

The previous lemma does not make any assumption on the objective function. With regard 
to $\sphere^+$, it implies that the global best (assuming $G_0>0$) will be improved after
some time almost surely.
However, since the precondition is that the particle has not improved for a while, 
this is not yet sufficient to ensure
finite hitting time to an $\epsilon$-ball around the optimum. One 
might imagine  
that the global best position is constantly updated while its value converges 
to some value greater than~$0$. 

A closer look into the proofs of Lemma~\ref{thm:exp-xt} and 
Lemma~\ref{thm:var-xt} and the underlying difference equations 
in \cite{JiangLuoYangPSO} 
reveals that 
they also apply to every particle~$i$ where  $(G_t-P^{(i)}_t)^2$ 
converges to a fixed value. 
%
In fact, as we will show 
in Lemma~\ref{lem:convergence-global-best}, it holds that 
%
%
$(G_t-P^{(i)}_t)^2$ 
almost surely is 
a null sequence for every positively initialised particle if certain assumptions 
on the parameters are met. Informally, this means 
that the personal best converges to the global best on $\sphere^+$, 
which might be considered as a corrected version of the 
erroneous Statement~\ref{statement:jiang-theorem} in 
\cite{JiangLuoYangPSO}.

\begin{lemma}
\label{lem:convergence-null-sequence}
\label{lem:convergence-global-best}
Consider  the basic PSO on $\sphere^+$. 
If $f(1)>\max\{\varphi_1^2,\varphi_2^2\}(1+\omega)/6$ and 
the assumptions from Theorems~\ref{thm:exp-xt} and~\ref{thm:var-xt} hold, then 
$(G_t-P^{(i)}_t)^2$ is a null sequence for every particle~$i$ that 
satisfies $P_0^{(i)}\ge 0$.
The statement also holds for the Noisy PSO if additionally 
$f(1)>1/3$ is assumed.
\end{lemma}
\begin{proof}
If there is no particle satisfying $P_0^{(i)}\ge 0$, nothing is 
to show. Otherwise, we have $G_0\ge 0$. Pick an arbitrary 
particle $i$ satisfying $P_0^{(i)}\ge 0$ and 
assume that $(G_t-P^{(i)}_t)^2$ is not a null sequence for 
this particle. Because 
of the special properties of one-dimensional $\sphere^+$ and the conditions $G_0\ge 0$ and 
$P_0^{(i)}\ge 0$, the sequences $G_t$ $P_t^{(i)}$ are 
monotone
decreasing and bounded, hence they are convergent. 
Assume 
that $(G_t-P^{(i)}_t)^2$ converges to some non-zero value. 
According to Theorems~\ref{thm:exp-xt} and~\ref{thm:var-xt}, 
$E(X_t)$ and $\Var(X_t)$ converge, more precisely  
it holds for the expectation that $\lim G_t < \lim E(X_t) < \lim P^{(i)}_t$, 
where all limits (also in the following) are for $t\to\infty$. 
In the case of the basic PSO, we obtain from 
Theorems~\ref{thm:exp-xt} and~\ref{thm:var-xt} that 
$
\lim E(X_t) = (\varphi_1 \lim P^{(i)}_t + \varphi_2 \lim G_t)/(\varphi_1+\varphi_2)
$
and 
\begin{align*}
\lim \Var(X_t)  = \frac{1}{6}\frac{\varphi_1\varphi_2}{(\varphi_1+\varphi_2)^2}
(\lim G_t-\lim P^{(i)}_t)^2\frac{(1+\omega)}{f(1)}
.\end{align*}
If $f(1)>\varphi_2^2(1+\omega)/6$, we obtain 
$\lim \Var(X_t) < 
(\varphi_1/(\varphi_1+\varphi_2))^2(\lim G_t-\lim P^{(i)}_t)^2
= 
(\lim G_t-\lim E(X_t))^2$, and if 
$f(1)>\varphi_1^2(1+\omega)/6$ we obtain 
$\lim \Var(X_t) < (\lim P_t^{(i)}-\lim E(X_t))^2$. If 
both inequalities apply, then the variance 
is smaller than the smallest of the two squared 
distances, and Chebyshev's inequality
yields that 
$\card{X_t-\lim E(X_t)} < \min\{\card{\lim P_t^{(i)}-\lim E(X_t)},
\card{\lim P_t^{(i)}-\lim E(X_t)}\}$, implying  
$G_t < X_t < P_t^{(i)}$, will occur with positive probability 
for sufficiently large~$t$ (using the same 
methods as in the proof of Lemma~\ref{lem:improvement-deltatenth} 
the errors become negligible if~$t$ is large enough). This leads to an improvement 
of $P_t^{(i)}$ by a positive amount 
and also $(P_t^{(i)}-G_t)^2$ will decrease by a positive amount.
Note the lower bound on the size of the positive improvement 
does not change as time increases.
As $t$ approaches infinity, the improvement will happen  
almost surely.

In the case of the Noisy PSO, the argumentation 
is similar. However, since the limit of the variance 
increases by $(\delta^2/12)/f(1)$ according to 
Theorem~\ref{thm:var-xt}, we will decompose 
the stochastic process in the same 
way as in the proof of Lemma~\ref{lem:improvement-deltatenth} 
and combine the calculations that follow from 
$f(1)>1/3$ with the considerations presented above for 
the basic PSO.  
For the variable $Y_t$, Chebyshev's inequality 
yields that 
$\card{Y_t-\lim E(X_t)} < \card{\lim P_t^{(i)}-\lim E(X_t)}+0.4899\delta$
and $\card{Y_t-\lim E(X_t)} < \card{\lim G_t-\lim E(X_t)}+0.4899\delta$ 
both occur with positive probability. Hence, the support of 
$X_t=Y_t + \Delta$ is a superset of a subset of $[\lim G_t,\lim P_t^{(i)}]$
with positive measure. 
Therefore, an improvement by a certain positive amount has positive 
probability and will occur 
almost surely as $t$ tends to infinity.
\end{proof}

Remark: The preconditions are satisfied for $\omega=0.4$,
$\varphi_1=\varphi_2=1.5$, which is included in the convergence region
of \cite{JiangLuoYangPSO} (c.f. Figure \ref{fig:convergence-regions}).
The proof of the lemma is in the appendix.
We can formulate the announced generalisation of 
Lemma~\ref{lem:improvement-deltatenth}. 

\begin{lemma}
\label{lem:improvement-deltatenth-all}
  Assume that
  $f(1)>1/3$ and
  that the conditions in Theorem \ref{thm:exp-xt} and Theorem
  \ref{thm:var-xt} hold. Consider the noisy PSO on $\sphere^+$, pick a particle~$i$ satisfying $P_0^{(i )}\ge 0$ 
   and denote  
   $G=\lim_{t\to\infty} G_t$.  
  Then for all sufficiently small 
  $\varepsilon'>0,$ there exists a
  $t_0>0$ such that $\forall t\geq t_0$, 
$\prob{G^*-\delta \leq X_t\leq G^*-\frac{\delta}{100}+\varepsilon'}\geq 3/100000$,
  where $X_t$ is the position of the considered particle in iteration $t$.
\end{lemma}

\begin{proof}
By Lemma~\ref{lem:convergence-null-sequence}
$(G_t-P^{(i)}_t)^2$ converges to~$0$ for particle~$i$. The 
stability analysis of the inhomogeneous difference equations for 
$\expect{X^{(i)}_t}$ and $\var{X_t^{(i)}}$ 
does not change 
and 
we still get 
$
    \tlim \expect{X_t^{(i)}} = G \text { and }  \tlim \var{X_t^{(i)}} = \delta^2/12f(1)
$ 
  as in 
Lemma~\ref{lem:improvement-deltatenth}. 
  The proof is completed as before.
\end{proof}

We are ready to prove the main result in this section.

\begin{proof}[Proof of Theorem~\ref{theo:main-theo-noisy}]
Since it is monotonically decreasing and bounded, the sequence 
$G_t$ has a limit~$G$. If $G$ is in the $\epsilon$\nobreakdash-ball 
around the origin, nothing is to show. Otherwise, we have 
$G>\epsilon$ and 
according 
to Lemma~\ref{lem:improvement-deltatenth-all}, some point 
$X\in[G-\delta,G-\delta/100]$ will be sampled almost surely in finite time. 
After a finite number of such improvements, the $\epsilon$-ball 
around the optimum will have been reached.
\end{proof}

\section{Conclusions}

Much of the theoretical research on the particle swarm optimiser has
focused on its convergence properties. In particular, conditions have
been found which has been claimed to guarantee mean square
convergence.  We point out an error in the proof of this claim,
showing that the mean square convergence property does not hold for
all functions. Still, we think particle convergence is not always
desirable, in particular when it occurs in non-optimal points in the
search space. To better understand the PSO as an optimiser, we suggest
to put more effort in understanding the expected first hitting time
(FHT) of the algorithm to an arbitrarily small $\epsilon$-ball around
the optimum.  We point out non-trivial configurations where the
basic PSO has infinite expected FHT on even simple problems like the \sphere
function. As a remedy to this undesirable situation, we abandon
convergence in mean square, and propose the Noisy PSO which has
non-zero particle variance, but finite expected FHT on the
one-dimensional \sphere function. 

\paragraph{Acknowledgement} The authors thank Ming Yiang for helpful
discussions.


\bibliographystyle{plain}



\end{document}